\documentclass[10pt,journal,compsoc]{IEEEtran}

\usepackage{mathrsfs}
\usepackage{amsthm,amsmath,amssymb}
\usepackage{algorithm}
\usepackage{subfigure}
\usepackage{graphicx}
\usepackage[misc]{ifsym}
\usepackage{multirow}
\usepackage{url}
\usepackage{amsmath}
\usepackage{amsthm}
\usepackage{booktabs}
\usepackage{algorithm}
\usepackage{algorithmic}

\begin{document}

\title{Robust Motion Averaging under Maximum Correntropy Criterion}

\author{Jihua Zhu,
        Jie Hu,
        Huimin Lu,
        Badong Chen,
        and~Zhongyu Li
\IEEEcompsocitemizethanks{\IEEEcompsocthanksitem J. Zhu, J. Hu and Z. Li are with Lab of Vision Computing and Machine Learning, School of Software Engineering, Xi'an Jiaotong University, Xi'an 710049, P.R. China.\protect\\
E-mail: zhongyuli@xjtu.edu.cn
\IEEEcompsocthanksitem H. Lu is with Kyushu Institute of Technology, Kitakyushu, Japan.
\IEEEcompsocthanksitem B Chen is with School of Artificial Intelligence, Xi’an Jiaotong University, Xi'an 710049, China.}

\thanks{}}

\markboth{}%
{Shell \MakeLowercase{\textit{et al.}}: Bare Advanced Demo of IEEEtran.cls for IEEE Computer Society Journals}

\IEEEtitleabstractindextext{%
\begin{abstract}
Recently, the motion averaging method has been introduced as an effective means to solve the multi-view registration problem. This method aims to recover global motions from a set of relative motions, where the original method is sensitive to outliers due to using the Frobenius norm error in the optimization. Accordingly, this paper proposes a novel robust motion averaging method based on the maximum correntropy criterion (MCC). Specifically, the correntropy measure is used instead of utilizing Frobenius norm error to improve the robustness of motion averaging against outliers. According to the half-quadratic technique, the correntropy measure based optimization problem can be solved by the alternating minimization procedure, which includes operations of weight assignment and weighted motion averaging. Further, we design a selection strategy of adaptive kernel width to take advantage of correntropy. Experimental results on benchmark data sets illustrate that the new method has superior performance on accuracy and robustness for multi-view registration.
\end{abstract}

}

\maketitle

\IEEEdisplaynontitleabstractindextext

%
\IEEEpeerreviewmaketitle

\section{Introduction}
\IEEEoverridecommandlockouts
\IEEEPARstart{P}{oint} set registration is a fundamental and important technique in many domains, such as compute vision, robotics, and computer graphics, etc. For each range scan given in a set-centered frame, the registration goal is to find an optimal rigid transformation (global motion) and transform it into the reference coordinate frame. Due to the number of involved point sets, point set registration can be divided into pair-wise registration and multi-view registration problems. In the past few decades, lots of effective approaches have been proposed to solve the pair-wise registration problem. Among these approaches, the iterative closest point (ICP) algorithm~\cite{besl1992method} is one of the most popular methods. Based on this basic algorithm, many ICP variants~\cite{rusinkiewicz2001efficient} have been proposed to improve the performance of pair-wise registration in different perspectives. For convenience, we will use the term rigid transformation and motion interchangeable throughout this paper.

Different from pair-wise registration, the multi-view registration problem is more complex and has attracted less attention. In the literature, some approaches have been proposed to solve this difficult problem. For example, Chen et al. \cite{chen1992object} proposed the alignment-and-merging approach, which repeatedly aligns and merges two scans until all scans are merged into the whole model. This approach is straightforward but suffers from the error accumulation problem. To address this issue, Evangelidis et al. \cite{evangelidis2017joint} proposed the JRMPC approach, which assumes that all points are realizations of a unique Gaussian mixture model (GMM) and therefore casts the registration into clustering problem. Subsequently, the expectation maximization (EM) algorithm is utilized to estimate GMM parameters as well as all global motions for multi-view registration. This approach is time-consuming due the large number of parameters required to be estimated. Therefore, Zhu et al. \cite{zhu2019efficient} introduced the $k$-means algorithm to solve the multi-view registration problem. Compared with the JRMPC, the $k$-means based approach is more efficient and likely to obtain better registration results.

As these approaches estimate each global motion sequentially, they are more likely to be trapped into a local minimum, especially when the scan number is large. Therefore, Krishnan et al. \cite{krishnan2005global} proposed the optimization-on-a-manifold approach to simultaneously optimize all motions. To obtain desired results, it requires to establish accurate point correspondences for all scan pairs, which is very difficult in practice. Subsequently, Mateo et al. \cite{mateo2014bayesian} extended this approach under the Bayesian perspective, which views pair-wise correspondence as missing data and solves the registration problem by the EM algorithm. Although this approach can simultaneously optimize all global motions, it requires to compute a huge number of latent variables, which is time consuming.

For multi-view registration, another feasible solution is to recover global motions from a set of relative motions. To this end, Govindu \cite{govindu2004lie} proposed the motion averaging (MA) algorithm, which avoids averaging of motion in Lie groups but performs average in the Lie-algebra of the underlying motion representation. With an initial guess, global motions can be simultaneously recovered from a set of relative motions by MA algorithm, which was further extent to solve multi-view registration problem \cite{govindu2014averaging}.
Although these two algorithm is effective, it is sensitive to outliers due to utilizing Frobenius norm error in optimization. Govindu \cite{Govindu2016robust} combined graph-based sampling scheme and Random sample consensus (RANSAC) method to remove motion outliers. This approach is more robust, but the efficiency is seriously reduced with the increase of scan number.
For robot mapping, Grisetti et al. \cite{kummerle2general} proposed the general framework for graph optimization, called as G2O, which takes the same inputs as that of MA algorithm. Similar to MA algorithm, it is effective but sensitive to outliers.

Besides, Bourmaud et al. \cite{Bourmaud2016bayes} proposed Bayesian MA algorithm for robot mapping. It is more complex than the original MA and its performance is greatly affected by the assignment of a reasonable covariance to each relative motion, which is very difficult in real applications. Meanwhile, Arrigoni et al. \cite{arrigoni2016global} introduced the low-rank and sparse (LRS) matrix decomposition to solve multi-view registration, which concatenates all available relative motions into a large matrix and then decomposes it into one sparse matrix and one low-rank matrix. This approach can be viewed as another MA method and that is robust to outliers, but it requires more relative motions to achieve good registration. What's more, these methods treat each relative motion equally, and this will reduce the performance of registration. Accordingly, Guo et al. \cite{guo2018weighted} proposed weighted MA algorithm and Jin et al. \cite{jin2018multi} proposed weighted LRS algorithm, which can really improve the performance of multi-view registration with each relative motion assigned by a suitable weight, e.g. reliable motions assigned with high weights. However, it is difficult to manually assign a suitable weight to each relative motion.

Previous MA methods often use Frobenius norm error in optimization, and they perform well under the assumption of Gaussian noises. However, in practice, a relative motion set often includes outliers. In this case, the Frobenius norm error can not properly capture error statistics, which may seriously degrade the performance. Recently, correntropy \cite{Principe2010} has been proposed as an information theoretic learning measure to solve robust learning problems \cite{chen2017kf,chen2016generalized,du2018robust,he2019robust}. Compared with Frobenius norm, correntropy includes all even moments of the error. Therefore, the correntropy measure is robust against outliers and can achieve better learning performance especially when data contain large outliers.

Accordingly, this paper introduces the correntropy measure to reformulate the MA problem, which is difficult to be solved directly. To this end, the half-quadratic (HQ) \cite{Nikolova2007hq} technique is utilized to transform the problem into a half-quadratic optimization problem, which can be solved by the traditional optimization method. Further, we design an adaptive selection strategy for kernel width to take advantage of correntropy properties. Compared with Frobenius norm error, the negative effects of outliers are therefore alleviated by the correntropy measure. In summary, the main contributions of this paper are delivered as
1)	It proposes a novel cost function for robust motion averaging.
2)	It develops an effective MA algorithm by the HQ technique.
3)	Experiments carried out on benchmark data sets confirm its superior performance over other state-of-the-art algorithms.

The remainder of the paper is organized as follows. Section 2 briefly briefly reviews the concepts of MCC and HQ optimization theory. Section 3 formulates the correntropy based objective function for motion averaging and proposes the HQ based algorithm. Following that is section 4, in which the proposed approach is tested and evaluated on four benchmark data sets. Finally, conclusions are drawn in Section 5.

\section{Preliminaries}

This section briefly reviews MCC and HQ optimization theory, which are bases of the proposed approach.
\subsection{Maximum correntropy criterion}
Given two random variables $X$ and $Y$, the correntropy is defined by:
\begin{equation}
V(X,Y) = E(\kappa (X,Y)) = \int {\kappa (x,y)} d{F_{XY}}(x,y)
\end{equation}
where $\kappa ( \cdot , \cdot )$ denotes a shift-invariant Mercer kernel and ${F_{XY}}(x,y)$ is the joint probability distribution function (PDF) of $(X,Y)$. In practice, the joint PDF is unknown and only a finite number of data points are available. With finite samples $\{ {x_i},{y_i}\} _{i = 1}^N$, the correntropy can be approximated as:
\begin{equation}
\hat V = \frac{1}{N}\sum\limits_{i = 1}^N {\kappa ({x_i},{y_i})}.
\end{equation}
Usually, the correntropy kernel utilizes Gaussian Kernel:
\begin{equation}
\kappa (x,y) = {G_\sigma }(e) = \exp ( - \frac{{{e^2}}}{{2{\sigma ^2}}}),
\end{equation}
where $\sigma$ is the kernel width and $e = (x - y)$ is the error term.

Obviously, the correntropy is a local and nonlinear similarity measure between two random variables within a "window" in the joint space defined by the kernel width. Compared with traditional measures, the correntropy contains all the even moments of the difference between $X$ and $Y$, and it is robust to outliers. In supervised learning, the correntropy measure based loss function is usually given by:
\begin{equation}
{J_{{G_\sigma }}} = \frac{1}{M}\sum\limits_{i = 1}^N {{\sigma ^2}} (1 - {G_\sigma }(e(i)))
\end{equation}
which is referred to as the MCC.

\subsection{Half-quadratic optimization theory}

Usually, it is difficult to directly optimize the correntropy based objective function, which is non-quadratic. Therefore, the HQ technique has been introduced to solve this problem.

According to the HQ theory \cite{Nikolova2007hq}, there is a convex conjugated function $\varphi$ corresponding to ${G_\sigma }(e)$ and they have the following relationship:
\begin{equation}
{G_\sigma }(e) = \mathop {\max }\limits_t (\frac{{{e^2}t}}{{{\sigma ^2}}} - \varphi (t)),
\label{eq:hq1}
\end{equation}
where  $t \in \mathbb{R}$ and the maximum is achieved at $t = {G_\sigma }(e)$. Equivalently, Eq. (\ref{eq:hq1}) can also be transformed into:
\begin{equation}
{\sigma ^2}(1 - {G_\sigma }(e)) = \mathop {\min }\limits_t (-{e^2}t + {\sigma ^2}\varphi (t)).
\label{eq:hq2}
\end{equation}
By defining $w =-t$ and $\phi (w) = {\sigma ^2}\varphi ( - w)$, Eq. (\ref{eq:hq2}) can be further derived as: \begin{equation}
\mathop {\min }\limits_e {\sigma ^2}(1 - {G_\sigma }(e)) = \mathop {\min }\limits_{e,w} ({e^2}w + \phi (w)).
\label{eq:hq3}
\end{equation}
Based on the HQ technique, the non-quadratic cost function is reformulated as the augmented objective function in enlarged parameter space $\{e,w\}$ by introducing auxiliary variable $w$.

\section{Robust Motion Averaging under MCC}
This section states the MA problem in multi-view registration and then proposes a robust solution under MCC.

\subsection{Problem statement}
Given multiple range scans, the goal of multi-view registration is to estimate the rigid transformation for each scan to the reference coordinate frame. For simplicity, the rigid transformation $({{\bf{R}}_i},{\vec t_i})$  can be defined in the form of motion ${\bf{M}}_i$ as:
\begin{equation}
{\bf{M}}_i = \left[ {\begin{array}{*{20}{c}}
{\bf{R}}_i&{\vec t}_i\\
0&1
\end{array}} \right].
\end{equation}
where ${{\bf{R}}_i}$ and ${\vec t_i}$ denote the rotation matrix and translate vector, respectively.
Compared with the multi-view registration problem, the pair-wise registration problem is much easier. Therefore, it is reasonable to achieve multi-view registration based on pair-wise registration, which arises the MA problem.
Given a set of estimated relative motions ${{\bf{\hat M}}_{ij}} \in \Omega $, it requires to recover the global motion $\{ {{\bf{M}}_i}\} _{i = 1}^N$ for multi-view registration. Accordingly, the multi-view registration can be formulated the following optimization problem:
\begin{equation}
\mathop {\arg \min }\limits_{{{\bf{M}}_i},{{\bf{M}}_j}} \sum\limits_{{{{\bf{\hat M}}}_{ij}} \in \Omega } {\left\| {{{{\bf{\hat M}}}_{ij}} - {{\bf{M}}_i}^{ - 1}{{\bf{M}}_j}} \right\|_F^2} .
\end{equation}
As either ${{\bf{M}}_i}$ or ${{\bf{M}}_j}$ denotes the variable of global motion, we only preserve ${{\bf{M}}_i}$ as the variable for the simplicity. This problem has been solved by the original MA algorithm \cite{govindu2004lie}, which is sensitive to outliers due to the application of Frobenius norm error in the optimization.

To improve the robustness, we introduce correntropy as the error measure and reformulate the multi-view registration problem as the following optimization problem:\
\begin{equation}
\mathop {\min }\limits_{{{\bf{M}}_i}} {J_{{G_\sigma }}}({{\bf{M}}_i}) = \sum\limits_{{{{\bf{\hat M}}}_{ij}} \in \Omega } {{\sigma ^2}(1 - {G_\sigma }({{\left\| {{{{\bf{\hat M}}}_{ij}} - {{\bf{M}}_i}^{ - 1}{{\bf{M}}_j}} \right\|}_F}))}.
\label{eq:Obj}
\end{equation}
Eq. (\ref{eq:Obj}) denotes a non-convex and non-quadratic cost function, which is difficult to be directly minimized by traditional methods. To this end, the HQ technique should be utilized to minimize this function.

\subsection{Optimization by the HQ theory}

As shown in Eq. (\ref{eq:hq3}), minimizing the correntropy measure based loss function in terms of $e$ equals to minimizing an augmented cost function in an enlarged parameter space $\{ e,w\} $. Accordingly, the correntropy measure based objective function can be further formulated as:
\begin{footnotesize}
\begin{equation}
{J_{{G_\sigma }}}({{\bf{M}}_i}) = \mathop {\min }\limits_{{w_{ij}}} \sum\limits_{{{{\bf{\hat M}}}_{ij}} \in \Omega } {\left[ {\left\| {{{{\bf{\hat M}}}_{ij}} - {{\bf{M}}_i}^{ - 1}{{\bf{M}}_j}} \right\|_F^2{w_{ij}} + \phi ({w_{ij}})} \right]}.
\end{equation}
\end{footnotesize}
Further, we can define the augmented cost function:
\begin{small}
\begin{equation}
{J_{HQ}}({{\bf{M}}_i},{w_{ij}}) = \sum\limits_{{{{\bf{\hat M}}}_{ij}} \in \Omega } {\left[ {\left\| {{{{\bf{\hat M}}}_{ij}} - {{\bf{M}}_i}^{ - 1}{{\bf{M}}_j}} \right\|_F^2{w_{ij}} + \phi ({w_{ij}})} \right]}
\label{eq:obj2}
\end{equation}
\end{small}
According to the HQ optimization theory, we obtain the equivalent relation as follows:
\begin{equation}
\mathop {\min }\limits_{{{\bf{M}}_i}} {J_{{G_\sigma }}}({{\bf{M}}_i}) = \mathop {\min }\limits_{{{\bf{M}}_i},{{\bf{W}}_{ij}}} {J_{HQ}}({{\bf{M}}_i},{w_{ij}}).
\end{equation}
This optimization problem can then be solved by the alternating minimization procedure as follows:

(1) Optimization of $w_{ij}$: According to Eq. (\ref{eq:hq1}) and Eq. (\ref{eq:hq3}), the minimum of the objective function ${J_{HQ}}$ is achieved by $w = {G_\sigma }(e)$ for given a certain $e$. Therefore, the optimal solution of $w_{ij}$ can be estimated for the fixed ${{\bf{M}}_i}$ as:
\begin{equation}
{w_{ij}} = {G_\sigma }({\left\| {{{{\bf{\hat M}}}_{ij}} - {{\bf{M}}_i}^{ - 1}{{\bf{M}}_j}} \right\|_F})
\label{eq:w}
\end{equation}
This procedure can be viewed as the weight assignment operation, which assigns different weights to each relative motion based on the residual motion error. According to the property of Gaussian function, a relative motion with small error will be assigned with a large weight, and vice versa. Different from previous methods, we do not manually estimate a weight for each relative motion, but automatically calculate them by the residual motion error. Therefore, suitable weight can be assigned to each relative motion due to properties of the correntropy measure.

(2) Optimization of ${\bf{M}}_{i}$: For the fixed $w_{ij}$, Eq. (\ref{eq:obj2}) is simplified into the following optimization problem:
\begin{equation}
{{\bf{M}}_i} = \mathop {\arg \min }\limits_{{{\bf{M}}_i}} \sum\limits_{{{{\bf{\hat M}}}_{ij}} \in \Omega } {{w_{ij}}\left\| {{{{\bf{\hat M}}}_{ij}} - {{\bf{M}}_i}^{ - 1}{{\bf{M}}_j}} \right\|_F^2}
\label{eq:wma}
\end{equation}
Eq. (\ref{eq:wma}) denotes the weighted MA problem, where the negative impact of outliers can be seriously reduced due to the small weight assigned by the first procedure. Since this problem can be solved by the variant of original MA algorithm, we present the solution without any provement.

\subsection{Weighted motion averaging}
Given the relative motion set $\{ {{\bf{\hat M}}_{ij,h}}\} _{h = 1}^H$, the motion averaging algorithm requires initial global motions $\{ {\bf{M}}_i^0\} _{i = 1}^N$ to achieve multi-view registration by iterations.
For one relative motion ${{\bf{\hat M}}_{ij,h}}$ and previous global motion $ \{ {\bf{M}}_i^{k - 1}\} _{i = 1}^N$,
the residual relative motion is defined as:
\begin{equation}
\begin{array}{l}
\Delta {{\bf{M}}_{ij}} = {\bf{M}}_i^{k - 1}{{{\bf{\hat M}}}_{ij}}{({\bf{M}}_j^{k - 1})^{ - 1}}\\
\qquad  \quad  = {(\Delta {{\bf{M}}_i})^{ - 1}}\Delta {{\bf{M}}_j}.
\end{array}
\label{eq:ma1}
\end{equation}
Eq. (\ref{eq:ma1}) can be converted into the equivalent formulation:
\begin{equation}
\Delta {{\bf{m}}_{ij,h}} = (\Delta {{\bf{m}}_{j,h}} - \Delta {{\bf{m}}_{i,h}})
\label{eq:ma2}
\end{equation}
where $\Delta {\bf{m}} = \log (\Delta {\bf{M}})$. Subsequently, the function $vec()$ is utilized to extract parameters from $\Delta {{\bf{m}}}$ to form a column wise vector $\Delta v$ and then Eq. (\ref{eq:ma2}) is transformed into the following form:
\begin{equation}
\Delta {v_{ij,h}} = (\Delta {v_{j,h}} - \Delta {v_{i,h}})
\label{eq:ma3}
\end{equation}
where $\Delta {v = vec(\Delta {\bf{m}}})$.

As each relative motion is assigned with a weight in our approach, the $6 \times 6N$  block-matrix ${{\bf{D}}_{ij,h}}$ is  constructed with the $i$th and $j$th block-elements filling with $ - {w_{ij,h}}{{\bf{I}}_{6 }}$ and ${w_{ij,h}}{{\bf{I}}_{6 }}$:
\begin{small}
\begin{equation}
{{\bf{D}}_{ij,h}} = \left[ {\begin{array}{*{20}{c}}
 \cdots &{ - w_{ij,h}^k{{\bf{I}}_{6 }}}& \cdots &{w_{ij,h}^k{{\bf{I}}_{6 }}}& \cdots
\end{array}} \right]
\label{eq:D}
\end{equation}
\end{small}
where ${{\bf{I}}_{6}}$ denotes the $6 \times 6$ identity matrix. According to Eq. (\ref{eq:ma3}), there exists the following relationship:
\begin{equation}
{{\bf{D}}_{ij,h}} \mathfrak{A}= w_{ij,h}^k\Delta {v_{ij,h}}
\label{eq:4}
\end{equation}
where $ \mathfrak{A}= \left[ {\begin{array}{*{20}{c}}
{\Delta {v_1};}&{\Delta {v_2};}& \cdots &{\Delta {v_N}}
\end{array}} \right]$. To refine global motions, Eq. (\ref{eq:4}) can be extended to the situation of many relative motions $\{{{\bf{\hat M}}_{ij,h}}\} _{h = 1}^H$ as follows:
\begin{equation}
{\bf{D}} \mathfrak{A}=  \Delta {{\rm{V}}_{ij}},
\end{equation}
where $\Delta {\bf{V}} = \left[ {\begin{array}{*{20}{c}}
{w_{ij,1}^k\Delta {v_{ij,1}};}& \cdots &{w_{ij,H}^k\Delta {v_{ij,H}}}
\end{array}} \right] $ and ${\bf{D}} = \left[ {\begin{array}{*{20}{c}}
{{{\bf{D}}_{ij,1}};}&{ \cdots ;}&{{{\bf{D}}_{ij,H}}}
\end{array}} \right]$
. This formulation leads to vector $\mathfrak{A}$ including parameters of all residual global motions:
\begin{equation}
\mathfrak{A}={{\bf{D}}^\dag } \Delta {{\rm{V}}_{ij}}
\end{equation}
where ${{\bf{D}}^\dag }$ is the pseudo-inverse matrix of ${\bf{D}}$. Finally, elements of $\mathfrak{A}$ can be utilized to update each global motion as:
\begin{equation}
{\bf{M}}_i^k = {\bf{M}}_i^{k - 1}\exp (\Delta {{\bf{m}}_i})
\end{equation}
where $
\Delta {{\bf{m}}_i} = rvec(\Delta {v_i})
$ is residual global motions and $rvec()$ denotes the inverse function of $vec()$.

\subsection{Implementation}
Obviously, our method is local convergent. To obtain desired results, initial guess should be provided for global motions in advance. Besides, its performance is affected by the kernel width $\sigma$ in correntropy measure. In the literature, lots of works have illustrated that relatively large kernel width can offer high convergence speed but suffer from less accuracy, and vice versa. As our approach achieves multi-view registration by iterations, it is better to use an adaptive kernel width. Specifically, the kernel width is set to be large at the beginning of the iteration and it should decrease with the increase of the iteration number. As residual motion error of decreases with the increase of iteration number, it is reasonable to set the kernel width to be proportional with the residual error of all global motions, e.g. ${\sigma _{k}} = \alpha {e_{{\bf{M}},k}}$, where $\alpha$ is a preset parameter and ${e_{{\bf{M}},k}}$ denotes the residual motion error defined as:
\begin{equation}
{e_{{\bf{M}},k}} = {{\sum\nolimits_{h = 1}^H {{{\left\| {{{{\bf{\hat M}}}_{ij}} - {{({\bf{M}}_i^k)}^{ - 1}}{\bf{M}}_j^k} \right\|}_F}} } \mathord{\left/
 {\vphantom {{\sum\nolimits_{h = 1}^H {{{\left\| {{{{\bf{\hat M}}}_{ij}} - {{({\bf{M}}_i^k)}^{ - 1}}{\bf{M}}_j^k} \right\|}_F}} } H}} \right.
 \kern-\nulldelimiterspace} H}.
\label{eq:err}
\end{equation}
This setting can well balance the convergence speed and accuracy of the proposed method.

Based on the above description, the proposed method is summarized in Algorithm 1, where the parameter $\alpha$ will be discussed and determined in experiments.
\begin{algorithm}[tb]
\caption{Robust motion averaging under MCC}
\label{algorithm_SRLE}
\textbf{Input}: Initial global motions $\{ {\bf{M}}_i^0\} _{i = 1}^N$, \\
\hspace*{0.38in} relative motion set $\{ {{\bf{\hat M}}_{ij,h}}\} _{h = 1}^H$\\
\textbf{Output}: Global motions $\{ {{\bf{M}}_i}\} _{i = 1}^N$
\begin{algorithmic}[1] 
\STATE $k=0$;
\REPEAT
\STATE $k=k+1$;
\STATE  Compute the residual error ${e_{{\bf{M}},k}}$ by Eq. (\ref{eq:err});
\STATE Obtain the kernel width ${\sigma _{k }} = \alpha {e_{{\bf{M}},k }}$;
\STATE Calculate the weight $w_{ij,h}^k$ for ${\bf{M}}_{ij,h}^k$ by Eq. (\ref{eq:w});
\FOR{$(h = 1:H)$}
\STATE Generate ${\bf{D}}_{ij,h}$ by Eq. (\ref{eq:D});
\STATE Concatenate  ${{\bf{D}}_{ij,h}}$ into the matrix ${\bf{D}}$;
\STATE Stack $w_{ij,h}^k\Delta {v_{ij,h}}$ into the vector $\Delta {{\rm{V}}_{ij}}$ ;
\ENDFOR
\STATE $ \mathfrak{A}= {{\bf{D}}^\dag }\Delta {{\rm{V}}_{ij}}$;
\STATE $\forall i \in [2,N],\;{\bf{M}}_i^k = {\bf{M}}_i^{k - 1}\exp (\Delta {{\bf{m}}_i})$
\UNTIL{$({\bf{M}}'$s change is negligible) and $(k \ge K)$}
\end{algorithmic}
\label{Algorithm1}
\end{algorithm}

\begin{figure*}[t]
\centering
\subfigure[]{
\includegraphics[scale=0.25]{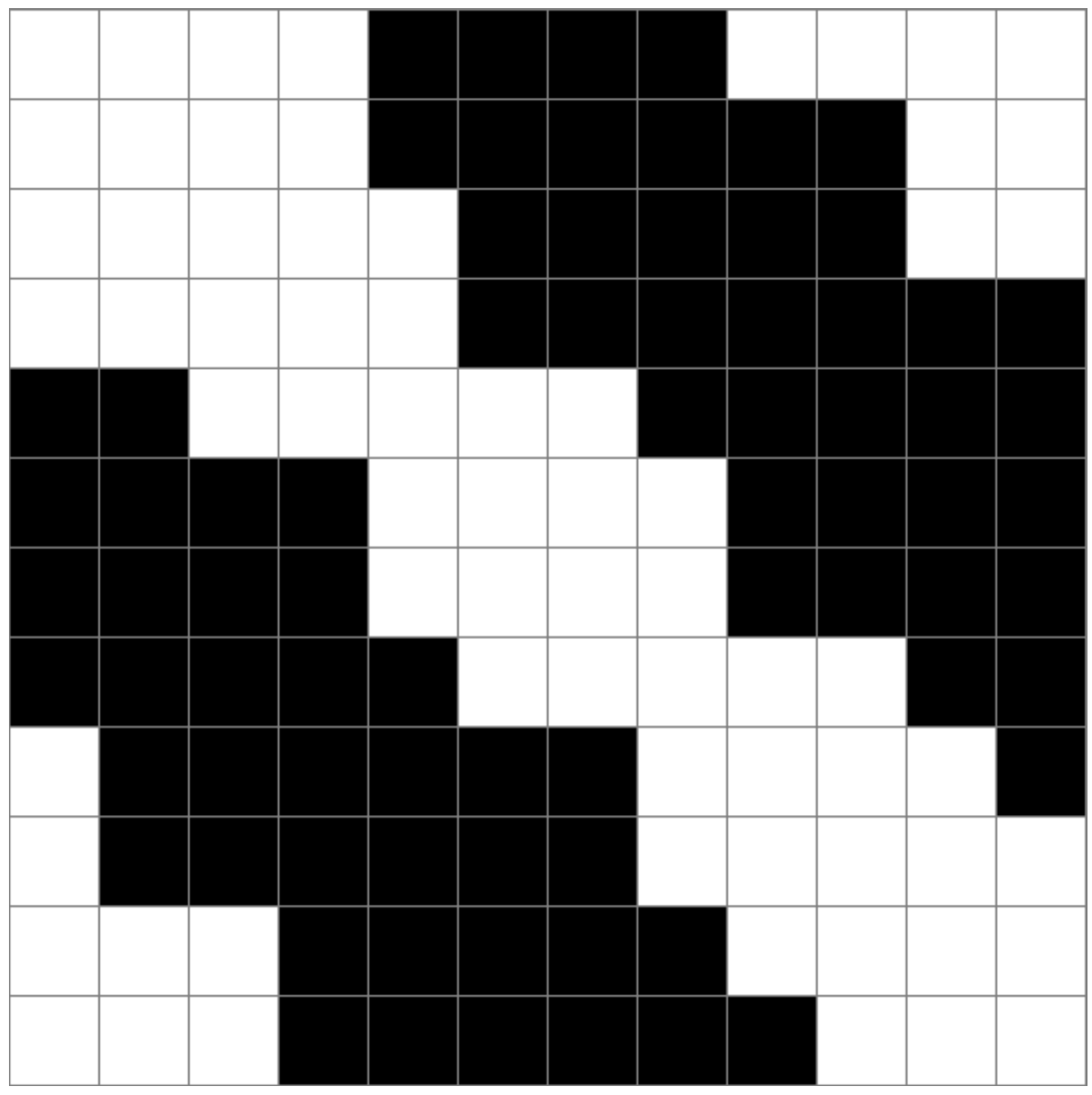}}
\subfigure[]{
\includegraphics[scale=0.25]{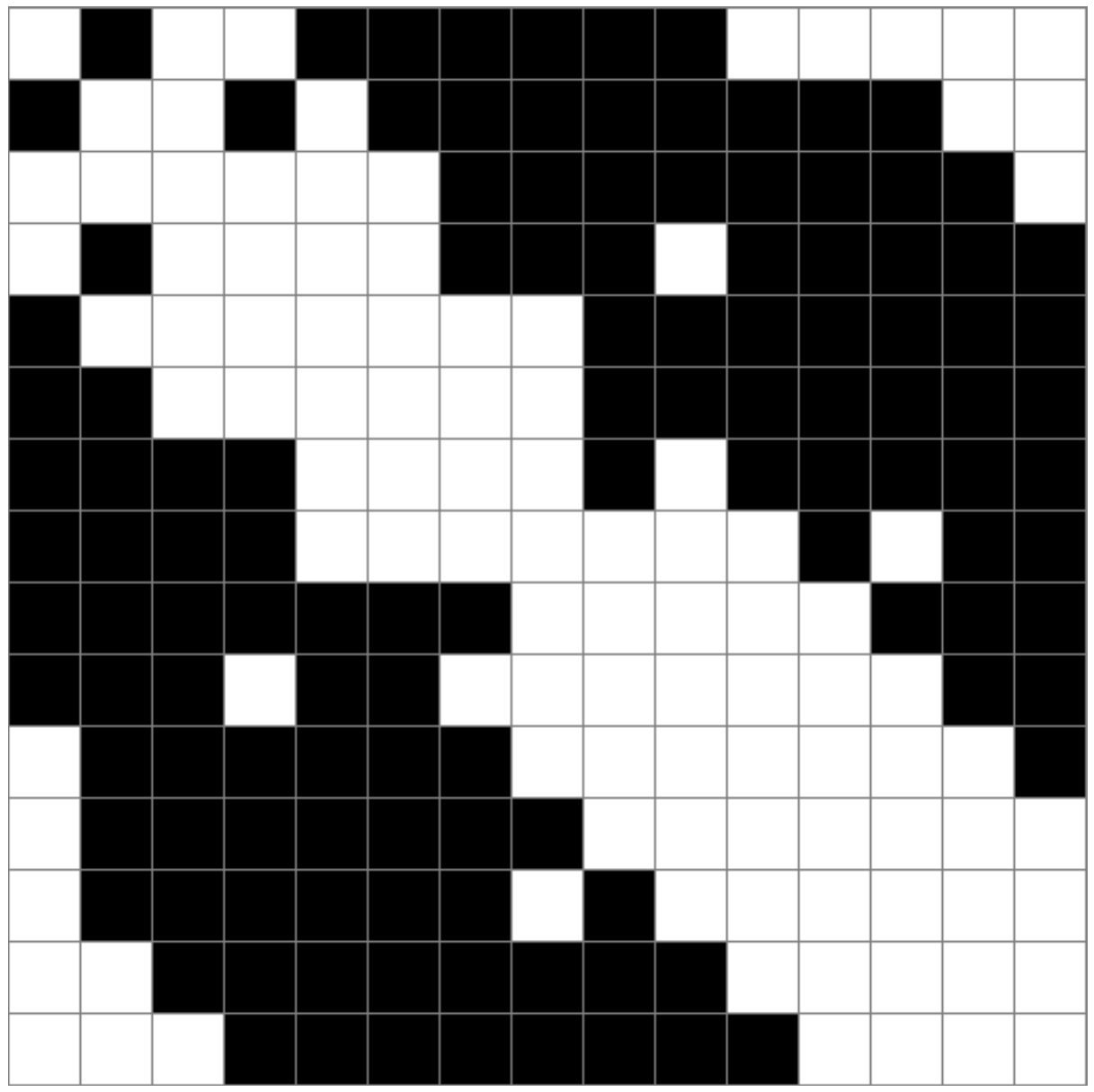}}
\subfigure[]{
\includegraphics[scale=0.25]{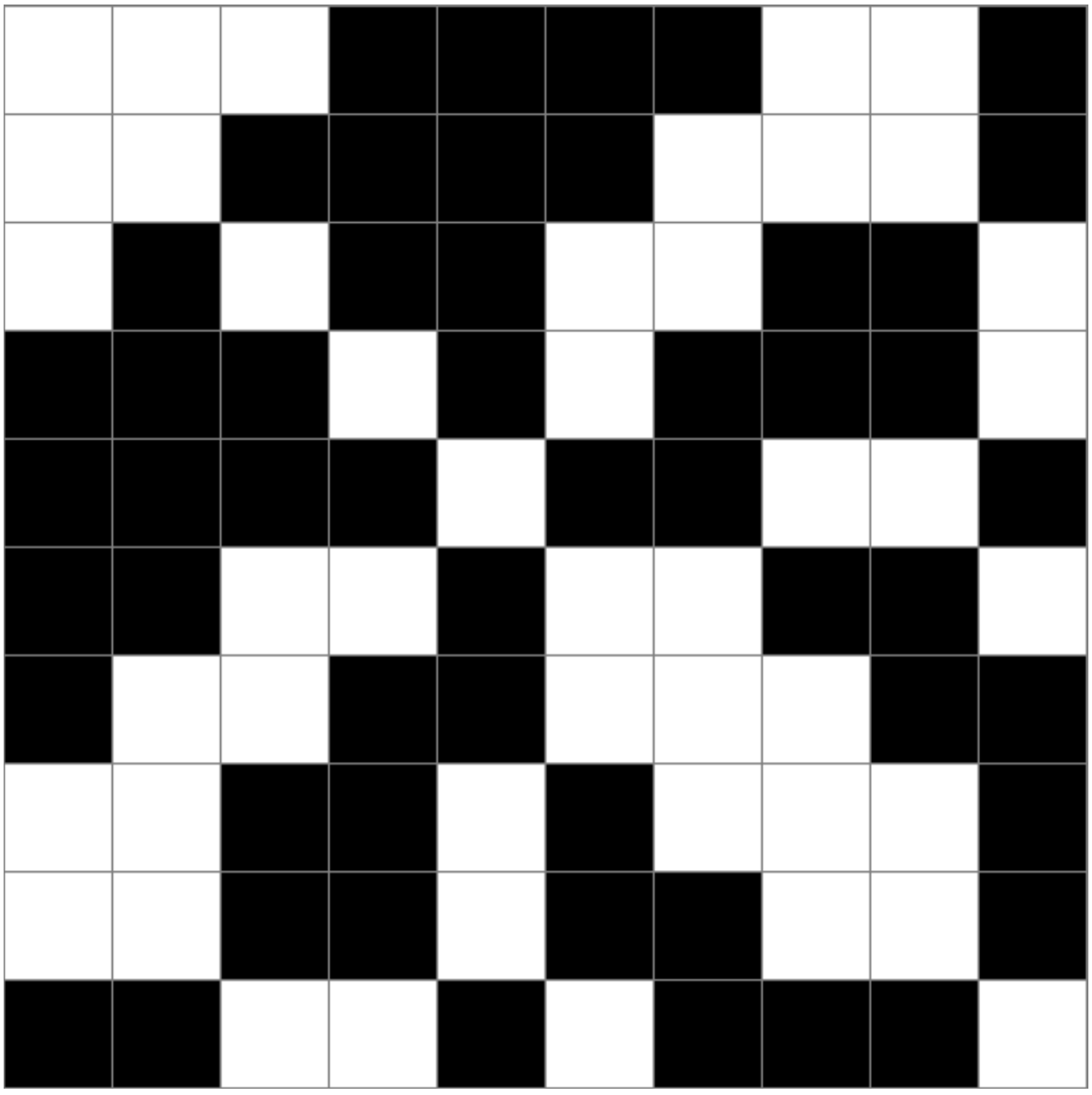}}
\subfigure[]{
\includegraphics[scale=0.25]{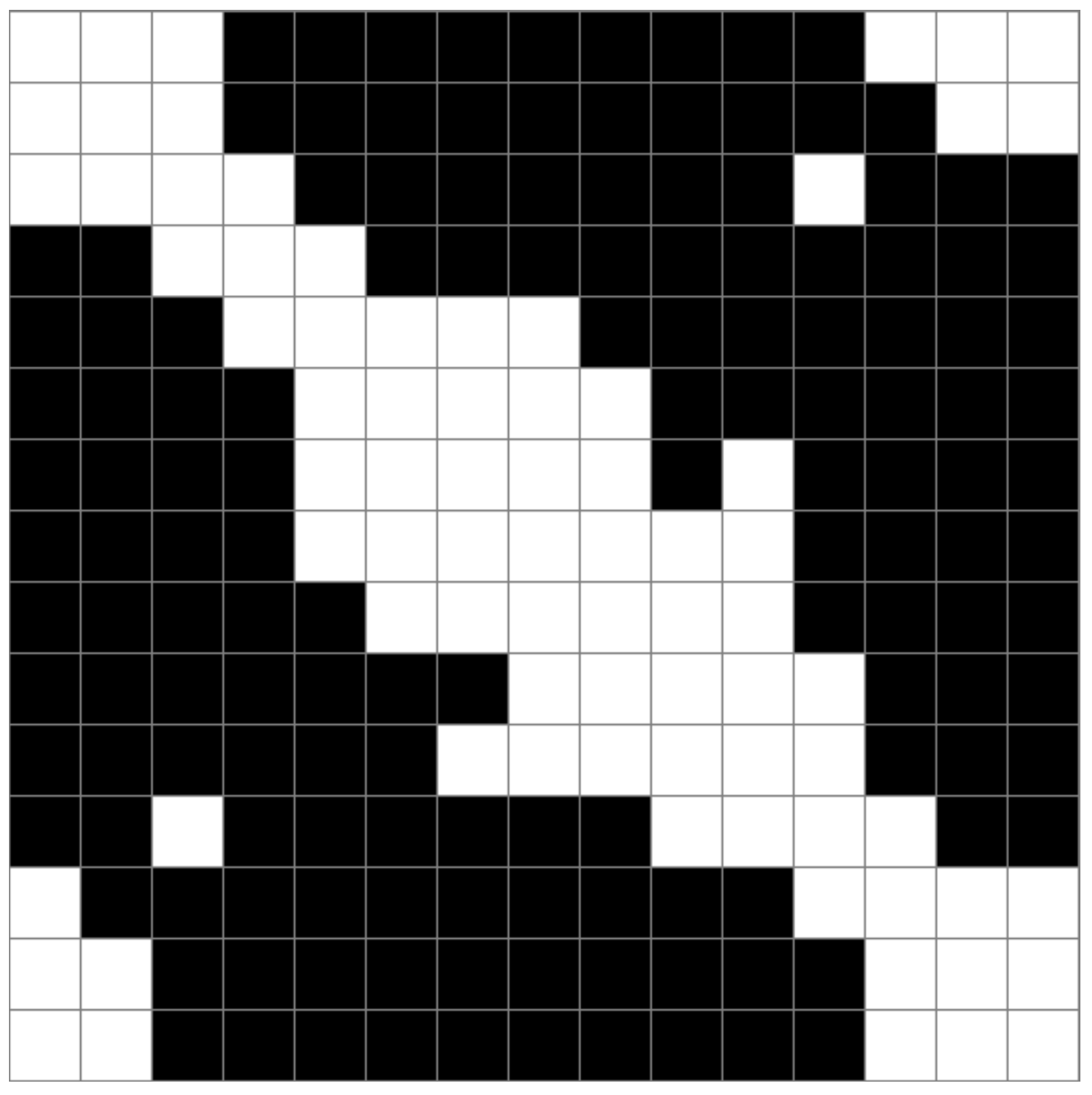}}
\subfigure[]{
\includegraphics[scale=0.21]{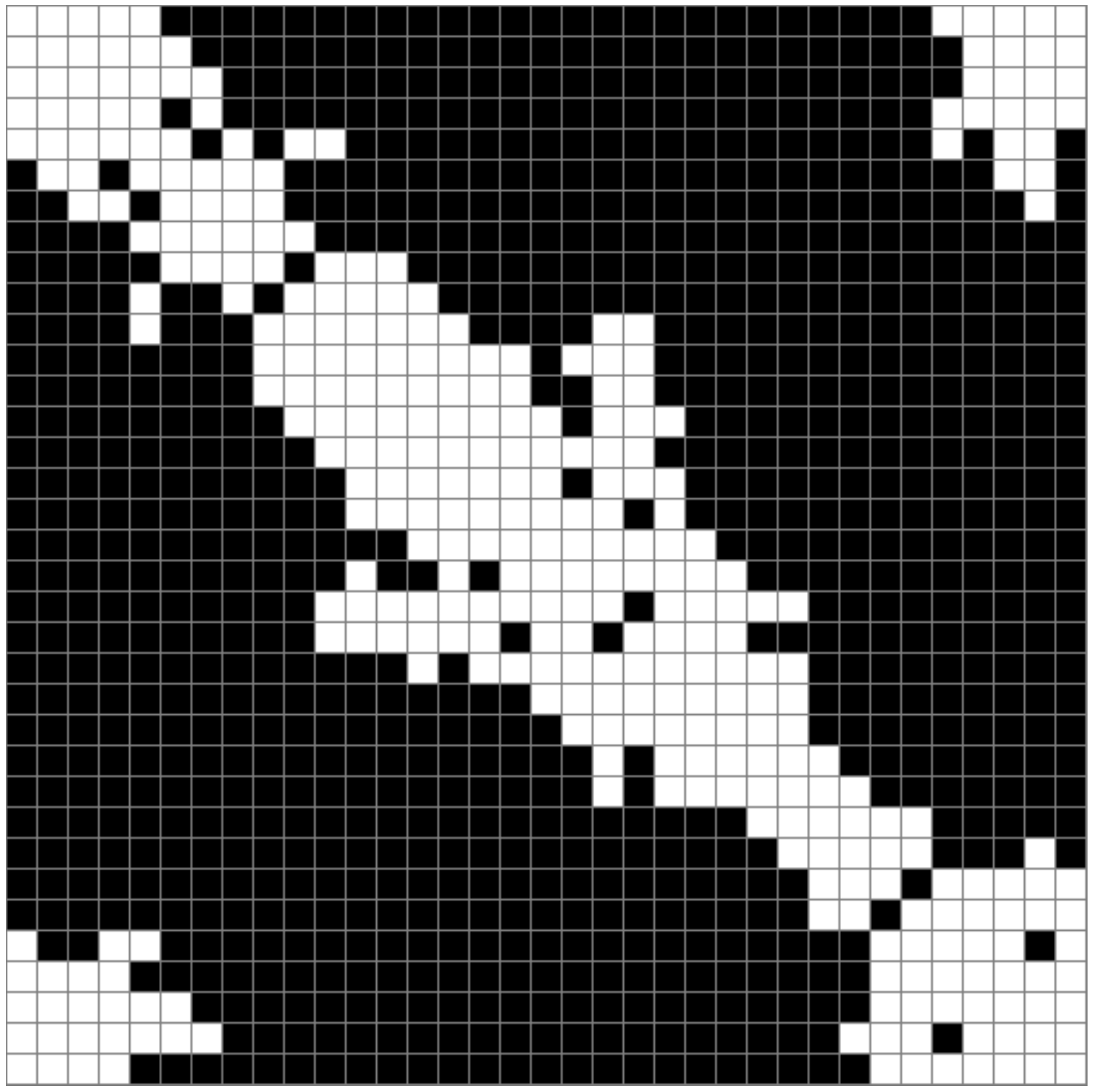}}
\caption{Demonstration of available relative motions in five data sets, where each small white square denotes one available motion and each small black square indicates one unavailable motion. (a) Stanford Armadillo with 47.22$\%$ available motions . (b) Stanford Buddha with 45.78$\%$ available motions. (c) Stanford Bunny with 46.00$\%$ relative motions available. (d) Stanford Dragon with 35.11$\%$ available motions. (e) Hand with 26.37$\%$ available motions}
\label{Fig:Data}
\end{figure*}

\section{Experiments}

\begin{table}[!htbp]
\centering
\caption{Details of benchmark datasets.}
\scalebox{0.85}{
\begin{tabular}{c|ccccc}
\midrule
Dataset  & \ Armadillo & \ Buddha & \ Bunny &\ Dragon &Hand\\
\midrule
Scan & 12 & 15 & 10 & 15 &36\\
Point & 307625 & 469193 & 362272   &1099005 &1605575\\
Motion & 68& 79 & 46 &103  & 323\\
\bottomrule
\end{tabular}}
\label{Tab:Data}
\end{table}

\begin{table}[!htbp]
\small
\centering
\caption{Statistics information of all relative motions for each data set.}
\scalebox{0.85}{
\begin{tabular}{l|ccc|ccc|}
\toprule
\multirow{2}{*} & \multicolumn{3}{c|}{Rotation error} & \multicolumn{3}{c|}{Translation error} \\
\cmidrule{2-4} \cmidrule{5-7}
&Mean	&Median	&RMSE &Mean	&Median	&RMSE\\
\midrule
Armadillo &	0.0059&	0.0041&	0.0054&	0.7365&	0.4666&	0.7052 \\
Buddha	&0.1984	&0.0124	&0.6150	&1.8631	&0.7141	&4.4277\\
Bunny	&0.0357	&0.0076	&0.0880	&0.0906	&2.4320	&6.1294\\
Dragon	&0.1368	&0.0061	&0.4930	&4.5063	&0.6799	&12.9122\\
Hand	&0.0103	&0.0026	&0.0625	&0.8366	&0.2411	&4.2900\\
\bottomrule
\end{tabular}}
\label{Tab:Sta}
\end{table}

This section tests and evaluates our approach on five benchmark data sets, where four data sets are taken from the Stanford 3D Scanning Repository \cite{turk1994zippered} and the Hand data set is provided by Torsello \cite{torsello2011multiview}. Each of them was acquired from one object model in multiple views and ground truth of rigid transformations was provided with multiple scans for the evaluation of registration results. But they are only utilized to assist for the final assessment. All experiments are performed on a four-core 3.6 GHz computer with 8 GB of memory.

\begin{figure*}[t]
\centering
\subfigure[]{
\includegraphics[scale=0.6]{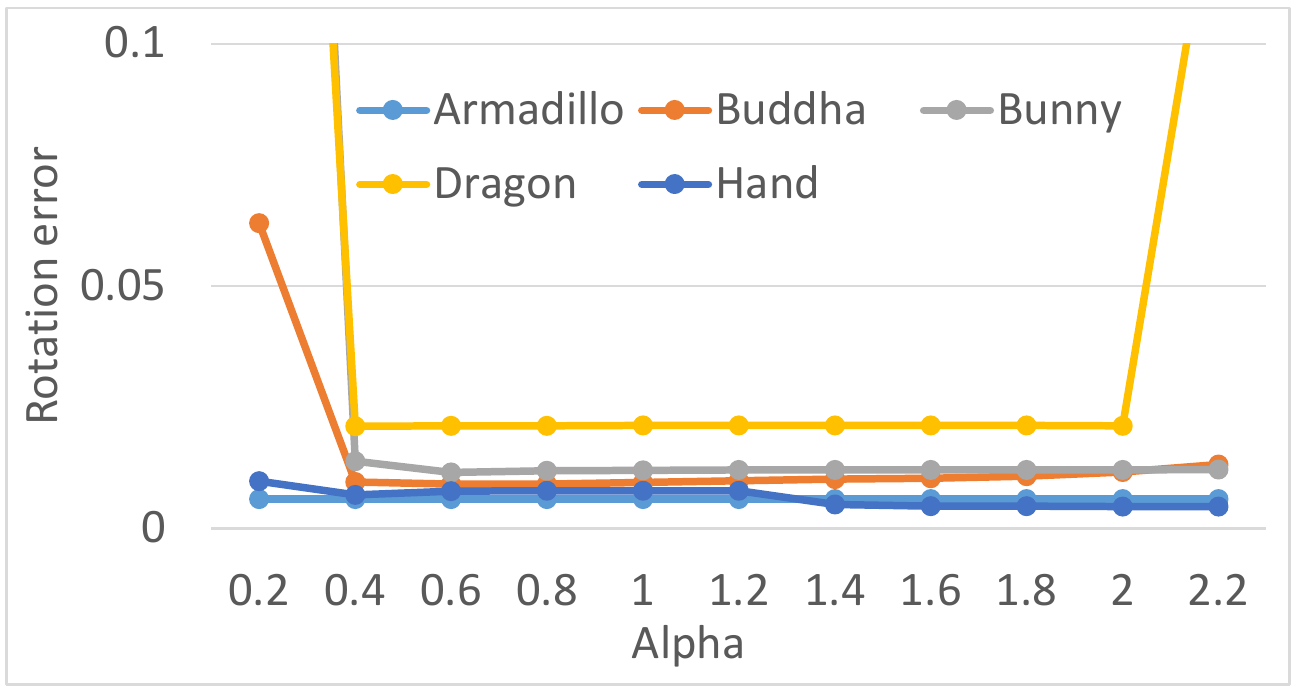}
}
\subfigure[]{
\includegraphics[scale=0.6]{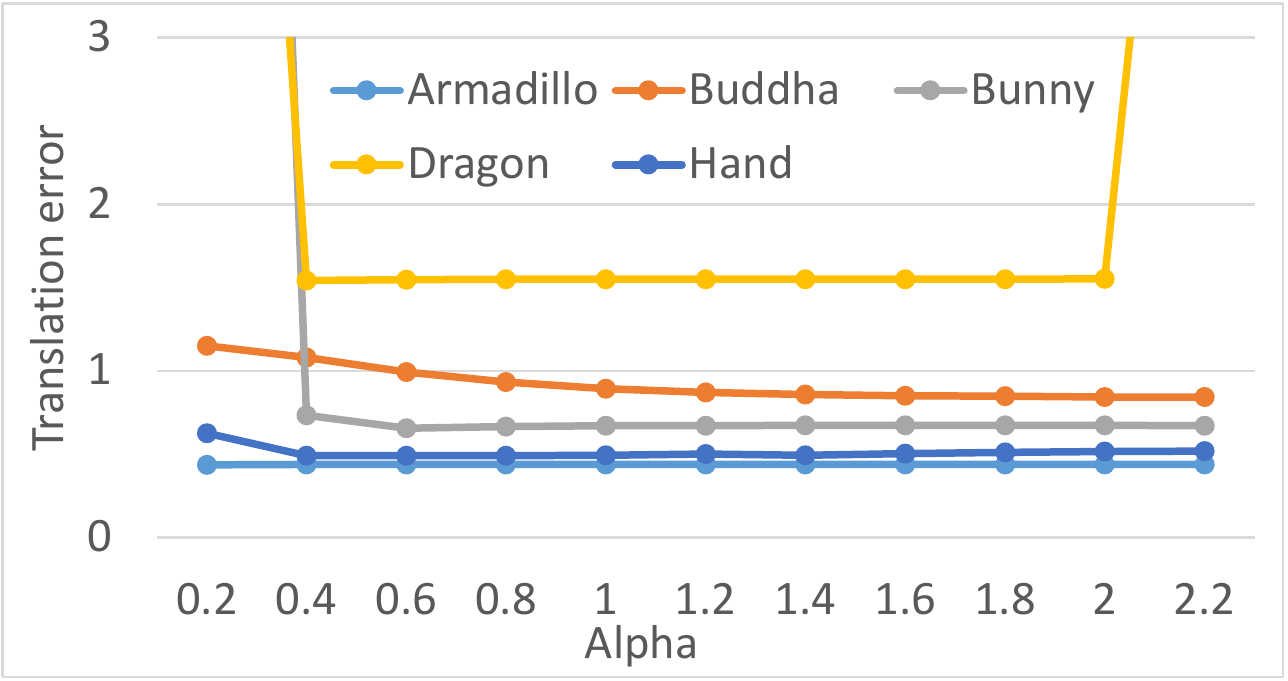}
}
\caption{Multi-view registration results of the proposed approach under varied $\alpha$ for five data sets. (a) Rotation error. (b) Translation error.}
\label{Fig:Para}
\end{figure*}

As the proposed approach takes relative motions as its input to recover global motions, we estimate relative motions for each scan pair by utilizing the pair-wise registration method proposed in \cite{lei2017fast}, which can obtain reliable results for these scan pairs with non-low overlap percentage. Given a set of scans, many scan pairs contain low overlap or non-overlapping percentages, their estimated relative motions are unreliable and meaningless. For accurate registration, it is better to utilize as many reliable relative motions but few unreliable relative motions as possible. Therefore, we only select relative motions of these scan pairs, whose trimmed mean square errors are less than the predefined threshold. For accuracy comparison, the registration error of rotation matrix and translation vector are defined as ${e_{\bf{R}}} = {\textstyle{1 \over M}}\sum\nolimits_{i = 1}^M {{{\left\| {{{\bf{R}}_{m,i}} - {{\bf{R}}_{g,i}}} \right\|}_F}} $
and ${e_t} = {\textstyle{1 \over M}}\sum\nolimits_{i = 1}^M {{{\left\| {{t_{m,i}} - {t_{g,i}}} \right\|}_F}} $, respectively. Here, $({{\bf{R}}_{g,i}},{t_{g,i}})$ indicates the ground truth of the $i$ th rigid transformation and $({{\bf{R}}_{m,i}},{t_{m,i}})$ denotes the one estimated by multi-view registration approach. Table \ref{Tab:Data} and Fig. \ref{Fig:Data} demonstrate some details of these four data sets as well as preserved motion sets. Besides, Table \ref{Tab:Sta} we list statistics information of all relative motions in each data set, including the mean, median, RMSE of rotation and translation errors. As shown in Table \ref{Tab:Sta}, each preserved motion set still contains outliers.

\subsection{Parameter tuning}
The performance of our approach is related to the selection of kernel width $\sigma$, which is set to be ${\sigma _{k}} = \alpha {e_{{\bf{M}},k}}$.
Empirically, we can set $\alpha=1$, which directly assigns the residual motion error to the kernel width. Here, we do experiments to find its appropriate value and check whether the performance of our method is sensitive to this parameter. More specially, we change the value of $\alpha$ in our approach and test it on all four data sets. Experimental results are reported in the form of registration errors $({e_{\bf{R}}},{e_t})$. During the experiment, we find the setting of $\alpha$ around 1.0 is more likely to obtain promising results for multi-view registration. Fig. \ref{Fig:Para} records registration results of our approach with varied values of $\alpha$ around 1.0.

As shown in Fig. \ref{Fig:Para}, we can observed that: 1) the setting of $\alpha \in [0.4,2]$ tends to obtain the desired results. 2) The performance of our approach is stable as long as $\alpha$ is set to be within a certain value range, e.g. $\alpha \in [0.4,2]$. Accordingly, the proposed approach is robust to the parameter $\alpha$ as long as it is chosen from a reasonable range, which makes it easy to apply this approach without much effort for parameter tuning.  However, both too large $\alpha$ and too small $\alpha$ may result in undesired registration results. For large $\alpha$, the correntropy measure is difficult to discriminate outliers from all relative motions, so the proposed method is unable to recover accurate global motions from a set of relative motions including outliers. For small $\alpha$, inliers with small noises may be viewed as outliers, which also makes the proposed method be unable to obtain desired registration results. Considering all these factors, we set $\alpha=1$, i.e. ${\sigma _{k}} = {e_{{\bf{M}},k}}$ in the proposed method in the following experiments.

\begin{table*}[!htbp]
\small
\centering
\caption{Registration results of different approaches tested on four data sets, where the numbers in bold denote the best performance.}
\scalebox{0.85}{
\begin{tabular}{l|ccc|ccc|ccc|ccc|ccc|}
\toprule
\multirow{2}{*} & \multicolumn{3}{c|}{Armadillo} & \multicolumn{3}{c|}{Buddha} & \multicolumn{3}{c|}{Bunny} & \multicolumn{3}{c|}{Dragon} & \multicolumn{3}{c|}{Hand}\\
\cmidrule{2-4} \cmidrule{5-7}\cmidrule{8-10} \cmidrule{11-13} \cmidrule{14-16}
&${e_{\bf{R}}}$& ${e_{\vec t}}$ & T(s) &${e_{\bf{R}}}$& ${e_{\vec t}}$ & Time(s)&${e_{\bf{R}}}$& ${e_{\vec t}}$ & T(s)&${e_{\bf{R}}}$& ${e_{\vec t}}$ & T(s)&${e_{\bf{R}}}$& ${e_{\vec t}}$ & T(s)\\
\midrule
LRS &0.3223 &8.8161 &\textbf{0.0508} &0.1985 &1.8254 &\textbf{0.0814} &0.0649 &3.5322 &\textbf{0.0353} &0.2760 &12.0721 &\textbf{0.0722}  &0.0187	&1.3681	& \textbf{0.5960} \\
G2O &0.3223	&8.8161	&0.6108 &0.1985	&1.8254	&0.8414 &0.0649	&3.5322	&0.5004 & 0.2760	&12.0721	&0.6895 &0.0142	&0.8677	&2.5293  \\
MA &0.6448 &10.0962 &0.6842 &0.4554 &3.4936 &1.1046 &0.0738 &4.5743 &0.1276 &0.8135 &20.5170 &0.9533 &0.0196 &	1.6076	&9.3326 \\
wMA &0.1216 &2.9548 &0.5562 &0.0143 &1.2425 &0.8419 &0.0380 &2.2822 &0.1438 &0.2583 &13.7669 &0.4624 &0.0345 &	2.6428	&9.8012\\
Ours  &\textbf{0.0061} &\textbf{0.4393} &0.6523n &\textbf{0.0103} &\textbf{0.8571} &1.2072 &\textbf{0.0121} &\textbf{0.6740} &0.2784 &\textbf{0.0213} &\textbf{1.5517} &1.0528  &\textbf{0.0078} &\textbf{0.4967} &15.7183\\
\bottomrule
\end{tabular}}
\label{Tab:Com}
\end{table*}

\begin{figure*}[!t]
\begin{center}
\includegraphics[width=0.9\linewidth]{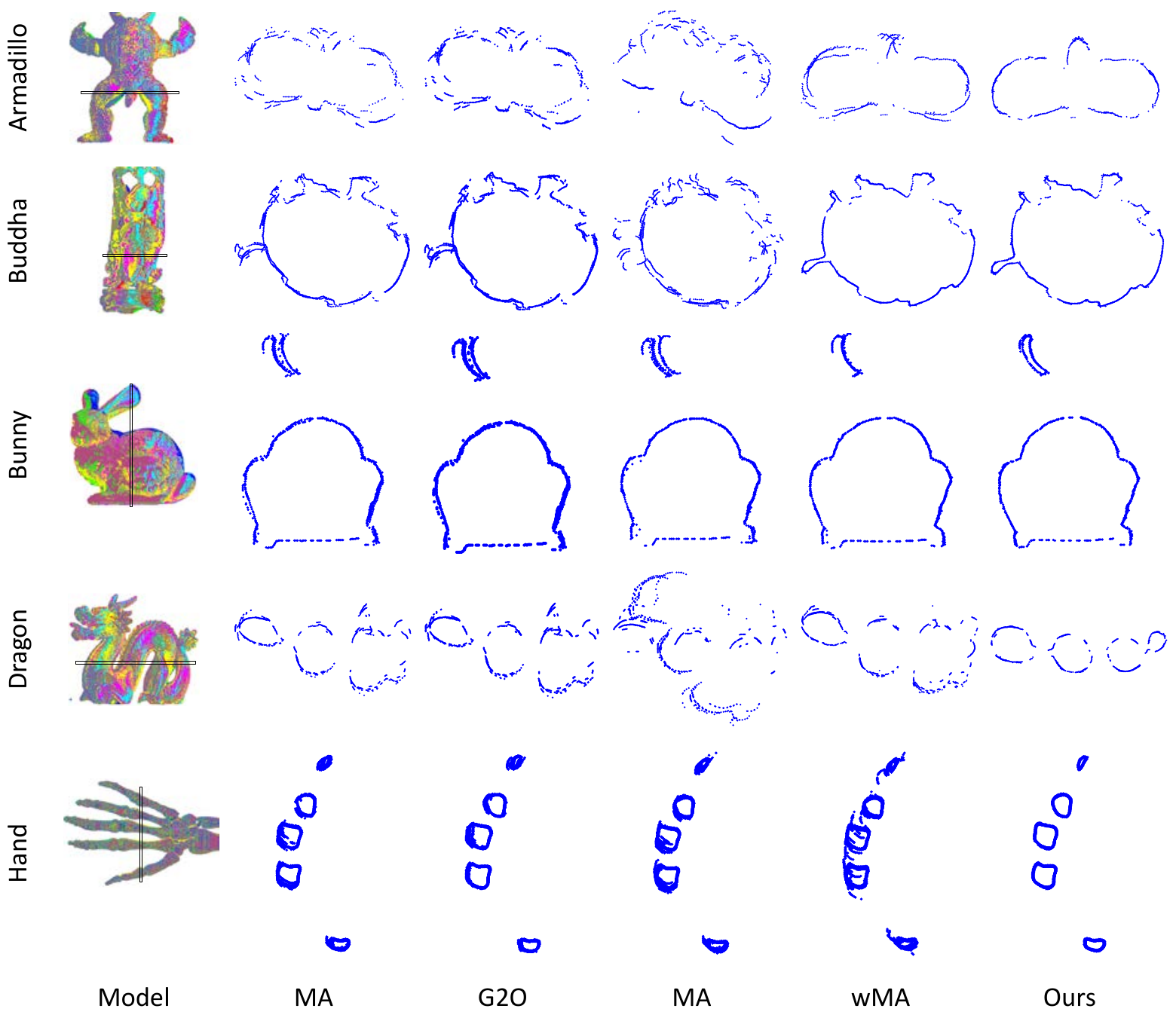}
\end{center}
   \caption{Multi-view registration results in the form of cross-section. (a) Aligned 3D models. (b) Results of LRS method. (c) Results of MA method. (d) Results of wMA method. (e) Our results.}
\label{Fig:Com}
\end{figure*}

\subsection{Results}

To demonstrate the performance, the proposed method is tested on four data sets and compared with some related approaches, including the multi-view registration approach based on the low-rank and sparse decomposition algorithm \cite{arrigoni2016global}, original motion averaging algorithm \cite{govindu2014averaging}, and weighted motion averaging algorithm \cite{guo2018weighted}, which are abbreviated as LRS, WA, and wWA, respectively. It should be noted that LRS does not requires initial guess for multi-view registration, but all other three approaches require initial global motions. As only relative motions are available in each data set, the output of LRS is taken as the input of other three approaches for multi-view registration. Experimental results are reported in the form of run time, rotation error, and translation error. These registration results are all recorded in Table \ref{Tab:Com}. For the evaluation of registration accuracy in a more intuitive manner, Fig. \ref{Fig:Com} displays all multi-view registration results in the form of a cross-section.

\begin{figure*}[t]
\centering
\subfigure[]{
\includegraphics[scale=0.5]{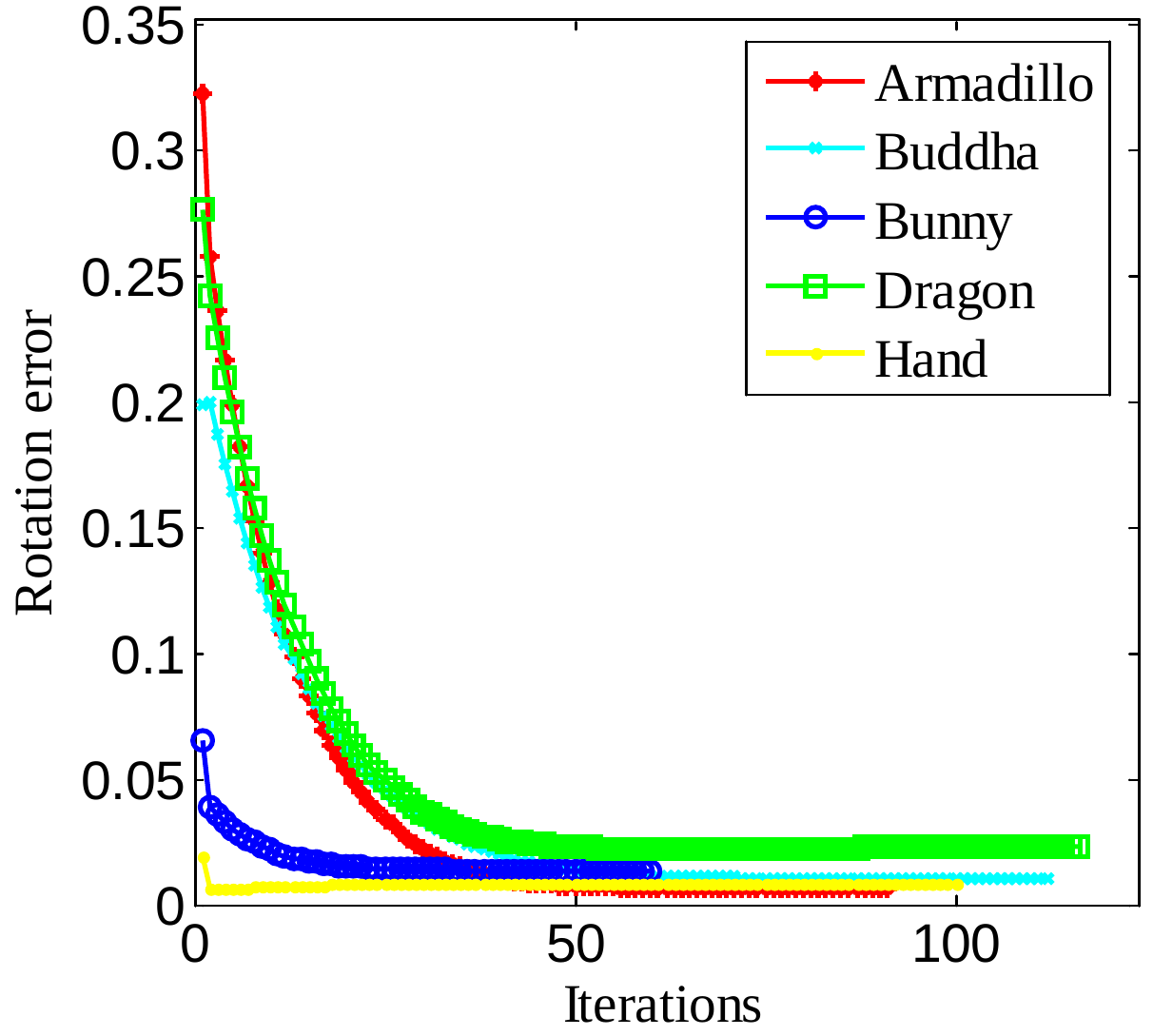}
}
\subfigure[]{
\includegraphics[scale=0.5]{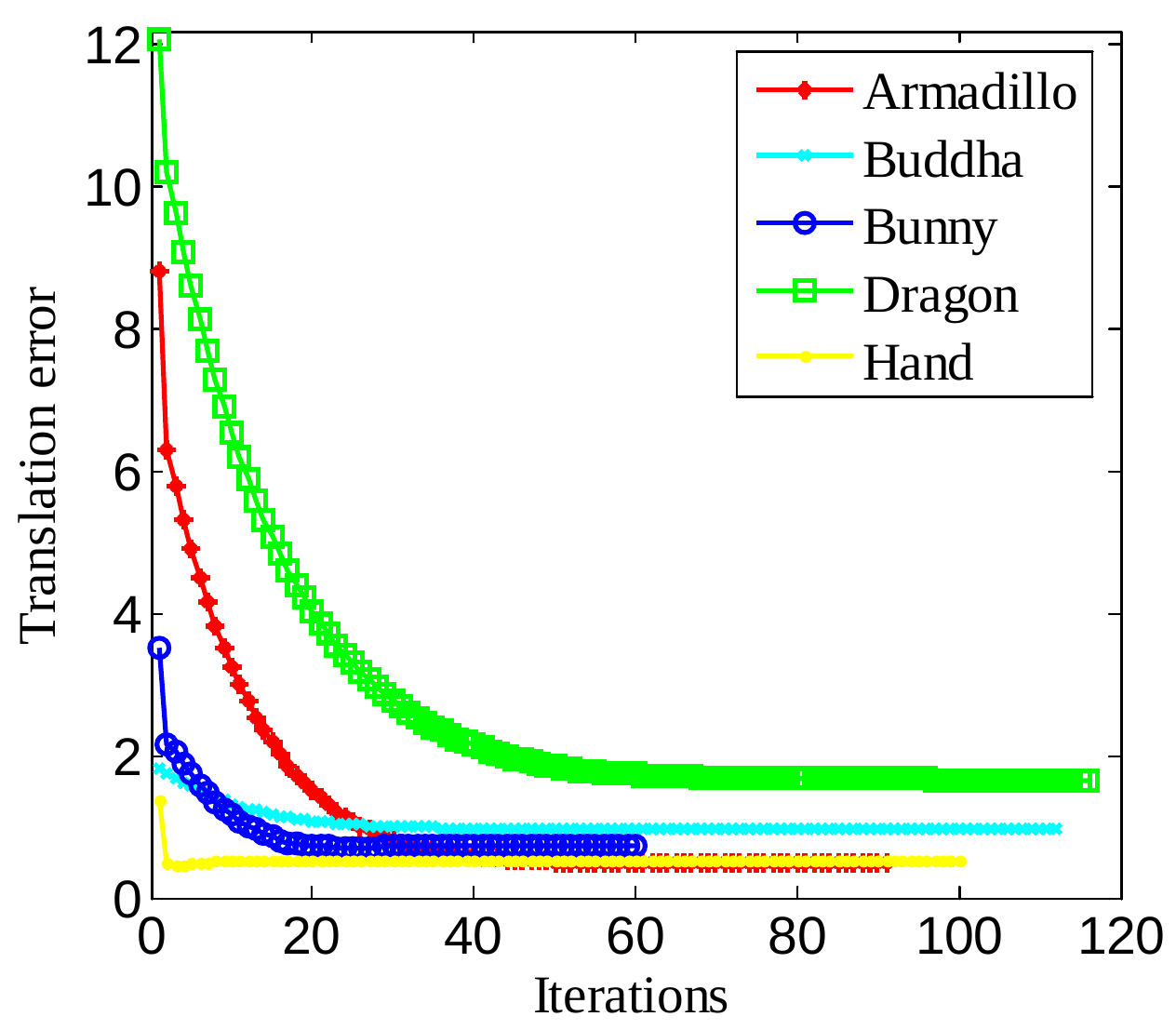}
}
\caption{Convergence curve of registration errors. (a) Rotation error. (b) Translation error.}
\label{Fig:Cur}
\end{figure*}

As shown in Table \ref{Tab:Com} and Fig. \ref{Fig:Com}, LRS is robust to unreliable relative motions. Without the initial guess of global motions, LRS may efficiently achieve multi-view registration by utilizing a set of relative motions. However, it requires a high proportion of available relative motions to obtain promising registration results. As shown in Fig. \ref{Fig:Data}, proportions of available relative motions are all below $50\%$ for these four data sets, so it is difficult to obtain promising registration results due to a low proportion of available relative motions.

Similar to robot mapping, the multi-view registration can be achieved by the G2O method, which takes relative motions and some results of LRS as its inputs. Besides, each relative motion requires to be assigned with one covariance matrix to denotes its uncertainty or reliability. Here, we assign the identity matrix to each relative motion due to the lack of prior information. As Table \ref{Tab:Com} and Fig. \ref{Fig:Com} demonstrate, the G2O method is efficient but is unable to obtain promising registration results. To obtain the desired results, each relative motion requires one appropriate covariance matrix, which is very difficult in most of practical applications.

For multi-view registration, both MA and wMA take registration results of LRS as their initial guess. As MA utilizes Frobenius norm error for the estimation of global motions, it is sensitive to outliers and difficult to obtain promising registration results due to the exiting of outliers. Different from MA, wMA pays more attention to reliable relative motions by assigning high weights. When each relative motion is assigned with one appropriate weight, e.g. outliers assigned with very low weight, wMA can obtain promising registration results, such as Stanford Buddha. However, the weight of each relative motion is estimated and assigned by some manual methods in wMA, it may assign a high weight to outliers, which can lead to the failure of multi-view registration.

Different from other competed methods, the proposed method utilizes the correntropy measure to achieve MA for muti-view registration. Compared with the Frobenius norm error, the correntropy measure can effectively alleviate the impact of large errors caused by outliers. For the balance of registration accuracy and convergence speed, adaptive kernel width has been selected by the well designed strategy. Therefore, the proposed method can achieve multi-view registration with promising results, even the input of relative motion set contains unreliable motions or outliers. The only weakness is that our method is less efficient than other competed methods due to weight calculation and a little more iterations.

To show its convergence, Fig. \ref{Fig:Cur} displays the curve of registration error for the proposed method. As shown in Fig. \ref{Fig:Cur}, even the inputs including outliers, the proposed method can can convergent to the desired registration results by iterations for the multi-view registration.

\section{ Conclusions}
In this paper, we proposed a novel and robust MA method for multi-view registration. To improve the robustness against outliers, it first utilizes the correntropy measure to design the objective function of MA, which arises a non-quadratic optimization problem. By the HQ theory, the correntropy based optimization problem can be solved by an alternating minimization procedure, which includes the operation of weight assignment and weighted MA derived from original MA algorithm.
Further, the selection strategy of adaptive kernel width is proposed to balance the accuracy and convergent speed of our algorithm. Experiments tested on benchmark data sets illustrate that the proposed approach can achieve multi-view registration with better performance than existing state-of-the art methods on accuracy and robustness. 

\section*{Acknowledgements}
This work is supported by the Fundamental Research Funds Central Universities; in part by State Key Laboratory of Rail Transit Engineering Informatization (FSDI) under Grant Nos. SKLKZ19-01 and SKLK19-09. We also would like to thank Andrea Torsello for providing Angel and Hand datasets.

\bibliographystyle{IEEEtran}
\bibliography{ref}

\end{document}